\documentclass[pdflatex,sn-mathphys-num, iicol]{sn-jnl}

\usepackage{graphicx}%
\usepackage{multirow}%
\usepackage{amsmath,amssymb,amsfonts}%
\usepackage{amsthm}%
\usepackage{mathrsfs}%
\usepackage[title]{appendix}%
\usepackage{xcolor}%
\usepackage{textcomp}%
\usepackage{manyfoot}%
\usepackage{booktabs}%
\usepackage{algorithm}%
\usepackage{algorithmicx}%
\usepackage{algpseudocode}%
\usepackage{listings}%
\usepackage{makecell}

\theoremstyle{thmstyleone}%
\theoremstyle{thmstyletwo}%

\theoremstyle{thmstylethree}%

\begin{document}

\title[Hierarchical Pooling and Explainability in GNNs for Tumor and Tissue-of-Origin Classification]{Hierarchical Pooling and Explainability in Graph Neural Networks for Tumor and Tissue-of-Origin Classification Using RNA-seq Data}


\author[1,2]{\fnm{Thomas} \sur{Vaitses Fontanari}}\email{tvfontanari@inf.ufrgs.br}

\author*[1,2]{\fnm{Mariana} \sur{Recamonde-Mendoza}}\email{mrmendoza@inf.ufrgs.br}


\affil*[1]{\orgdiv{Institute of Informatics}, \orgname{Universidade Federal do Rio Grande do Sul (UFRGS)}, \orgaddress{\city{Porto Alegre}, \state{RS}, \country{Brazil}}}

\affil[2]{\orgdiv{Bioinformatics Core}, \orgname{Hospital de Clínicas de Porto Alegre (HCPA)}, \orgaddress{\city{Porto Alegre}, \state{RS}, \country{Brazil}}}


\abstract{
This study explores the use of graph neural networks (GNNs) with hierarchical pooling and multiple convolution layers for cancer classification based on RNA-seq data. We combine gene expression data from The Cancer Genome Atlas (TCGA) with a precomputed STRING protein-protein interaction network to classify tissue origin and distinguish between normal and tumor samples. The model employs Chebyshev graph convolutions (K=2) and weighted pooling layers, aggregating gene clusters into 'supernodes' across multiple coarsening levels. This approach enables dimensionality reduction while preserving meaningful interactions. Saliency methods were applied to interpret the model by identifying key genes and biological processes relevant to cancer. Our findings reveal that increasing the number of convolution and pooling layers did not enhance classification performance. The highest F1-macro score (0.978) was achieved with a single pooling layer. However, adding more layers resulted in over-smoothing and performance degradation. However, the model proved highly interpretable through gradient methods, identifying known cancer-related genes and highlighting enriched biological processes, and its hierarchical structure can be used to develop new explainable architectures.  Overall, while deeper GNN architectures did not improve performance, the hierarchical pooling structure provided valuable insights into tumor biology, making GNNs a promising tool for cancer biomarker discovery and interpretation
}

\keywords{graph neural networks, pooling, gene expression data, cancer prediction}



\maketitle

\section{Introduction}\label{sec1}

Modeling gene expression data for genomic classification tasks is often challenging due to the limited number of samples compared to the high dimensionality of the data and the underlying non-linear patterns \citep{mostavi_cancersiamese_2021, hanczar2022assessment, SalvadorSanchez2018}.
Following the popularity of deep learning (DL) models in a diverse range of areas, many works have also started proposing neural networks (NNs) in the context of genomics, as an alternative to other machine learning (ML) strategies.
Moreover, NNs were also considered attractive due to their ability to automatically learn the relevant features and to represent non-linear relations \citep{hanczar2022assessment}.
Compared to traditional models, these approaches achieved state-of-the-art results in genomics classification tasks, provided a sufficient number of samples were available \cite{hanczar2022assessment}.

Recently, various works have proposed the use of graph neural networks (GNNs) \cite{rhee_hybrid_2017, ramirez_classification_2020, ramirez_prediction_2021, wang_single-cell_2021, yin_scgraph_2022} instead of fully-connected and convolutional neural networks (CNNs).
Their goal is to embed knowledge of gene interactions in the models in the hope that more accurate and robust classifiers could be produced.
The fundamental idea for justifying GNNs here is based on concepts coming from network medicine \citep{barabasi_network_2011}, where it is known that connected molecular elements in a biological network interact, and if an element of the network is related to a disease, it is more likely that its neighbors will also be.
Following that, diseases will be associated with localized clusters of genes in the biological network, and by mixing the expressions of close-by genes one could in principle produce more robust estimates of the expressions, and perhaps also identify local patterns in their interactions.

Previous work has focused mainly on the application of a single graph convolution and pooling layer.
Following that, the node embeddings were usually concatenated and passed through a fully connected network.
Some works also included additional modules in their frameworks, such as fully-connected networks parallel to the GNN and auto-encoders \citep{li_cancer_2021}.
A natural extension to the previous models is to increase the depth of these networks and to include forms of hierarchical pooling.
Introducing more convolutions and non-linearities could improve the modeling capacity of the model, while the hierarchical pooling scheme would better utilize the associations of nearby genes in the biological network in comparison with plain concatenation.
Our work provides an exploration of this idea on RNA-seq data, but, contrary to our expectations, we found that they did not improve the performance of the models, and the best results were still obtained using a single convolution layer or plain neural networks, depending on the task (Supplementary Material XYZ).

Furthermore, we also explore these models from the perspective of their explainability.
First, we show that using saliency methods developed for traditional CNNs also provided reasonable interpretations with the graph convolutions used here, revealing that the model used known cancer-related genes in the task of distinguishing tumor from non-tumor samples.

Secondly, we use the embeddings of the supernodes generated through hierarchical pooling to obtain higher-level explanations of the model.
In our scheme, each pooling step creates a supernode on a lower-dimensional graph and is associated with the original input nodes that were aggregated together to form that node.
We can analyze the cluster of nodes using over-representation analysis, which can provide insights into the biological processes most closely associated with the group of genes, along with other relevant information.
Then, saliency-based methods can be applied to the supernode embeddings to identify the most important ones for the classifier.
In this way, we were able to identify biological processes that were most relevant for a classification and a few of them were in fact associated with relevant biological processes in the tumor prediction task.

Therefore, the contributions of this work are summarized below:
\begin{enumerate}
    \item Contrary to our expectations, we demonstrate that increasing the number of graph convolutions and pooling layers don't lead to increased performance on RNA-seq tumor prediction and cohort classification tasks (Section \ref{sec:exp1}), and
    \item we show that these models can potentially be used for obtaining meaningful tumor marker genes, and to provide interpretations at the level of the biological processes (Section \ref{sec:exp2}).
\end{enumerate}

\section{Related Work}
Various approaches based on NNs 
have been proposed in recent years for genomic classification tasks, such as tumor/normal prediction, cancer cohort classification in pan-cancer scenarios, subtype classification, metastasis and other phenotypes \citep{joshi2022epicc, zhang_transformer_2022, divate2022deep, bourgeais_graphgonet_2022}.
\citet{hanczar2022assessment} and \citet{yu_architectures_2019}, for instance, have compared NNs with other traditional models for gene expression data, and obtained generally favorable results for the neural networks, as long as a sufficient number of examples is available.
Some works have also explored the use of Convolutional Neural Networks (CNN), in particular, for these tasks \citep{lyu2018deep, de2019deepgx, mostavi2020convolutional, shah_optimized_2020, liu_research_2022, chuang_convolutional_2021}.
In their work, \citet{lyu2018deep}, for example, argued that genes that are physically close together in the chromosome are more likely to interact.
Hence, by using CNNs, one could try to exploit the relations between adjacent genes.

Instead of assuming that close-by genes are more likely to interact, some works use graph convolutions over biological networks, whose connections explicitly represent the strength of the interactions \citep{rhee_hybrid_2017, ramirez_classification_2020, ramirez_prediction_2021, li_cancer_2021}.
One of the first studied GNNs was based on the ChebConv \citep{defferrard2016convolutional} and was developed for classifying breast cancer subtypes \citep{rhee_hybrid_2017}.
The model used the STRING PPI network as the backbone and RNA-seq data from TCGA.
Besides using a GNN, they incorporated also a Relation Network in the model \citep{santoro2017simple}.
Another GNN was proposed for pan-cancer classification of cancer cohort by \citet{ramirez_classification_2020}.
In their work, they explored the use of two backbone networks: one constructed from the STRING PPI network and another one constructed from a correlation matrix built using the gene expression data.
Later on, they also integrated clinical data with the GCN models for cancer survival prediction \citep{ramirez_prediction_2021}.

In general, these works have used only a single level of pooling, with the exception of \citep{rhee_hybrid_2017}, where 2 were used.
Some of the cited works have also studied ways in which their models could be interpreted to discover relevant biomarkers.
Particularly related to ours, \citet{mostavi2020convolutional} have used saliency maps to study a CNN for cohort classification, and were able to find genes known to be related to some types of cancers.

\section{Materials and Methods}

\subsection{Data}\label{sec:dataProcessing}
We obtained RNA-seq data for various types of cancer from The Cancer Genome Atlas (TCGA) \footnote{https://www.cancer.gov/ccg/research/genome-sequencing/tcga}.
Specifically, we downloaded the upper-quartile FPKM (UQ-FPKM) data from Xena \citep{goldman2020visualizing} for each of the 33 cancer cohorts available.
We retained only cohorts with at least 10 Primary Tumor and 10 Normal Tissue samples to ensure a sufficient number for training, validation, and testing.
Similarly to other works \citep{Duan2021, albaradei_metacancer_2021, chaudhary_deep_2018, wang_similarity_2014}, we removed genes and samples from each cohort that contained more than 20\% of missing values.
We used the log-transformed FPKM values, so that the features follow an approximate Gaussian distribution.

In our experiments with biological networks, we have generally used version 11.6 of the STRING network, restricted to the Homo sapiens \citep{szklarczyk2019string}.
We mapped each ENSG value from the TCGA data to their best matching ENSP identifier using the STRING API\footnote{https://string-db.org/cgi/help.pl?subpage=api}.
This reduced the number of features in the dataset significantly to 14148 genes.
Furthermore, this process also created 15 singleton nodes in the resulting graph, besides one big connected components containing all the remaining nodes.
Singleton nodes would cause problems for the clustering algorithms, and we have therefore chosen to drop these nodes, since they represented a very small percentage of the total network.
This resulted in a single connected component containing 14133 genes, each of which is associated with the expression value of a gene, here represented by the best matching ENSP identifier.
The final dataset contained 701 normal tissue samples and 7008 tumor samples.
The resulting number of examples in each TCGA pan-cancer dataset and of each type are summarized in the Supplementary Material.

\begin{figure*}[h]
\centering
\includegraphics[width=0.85\linewidth]{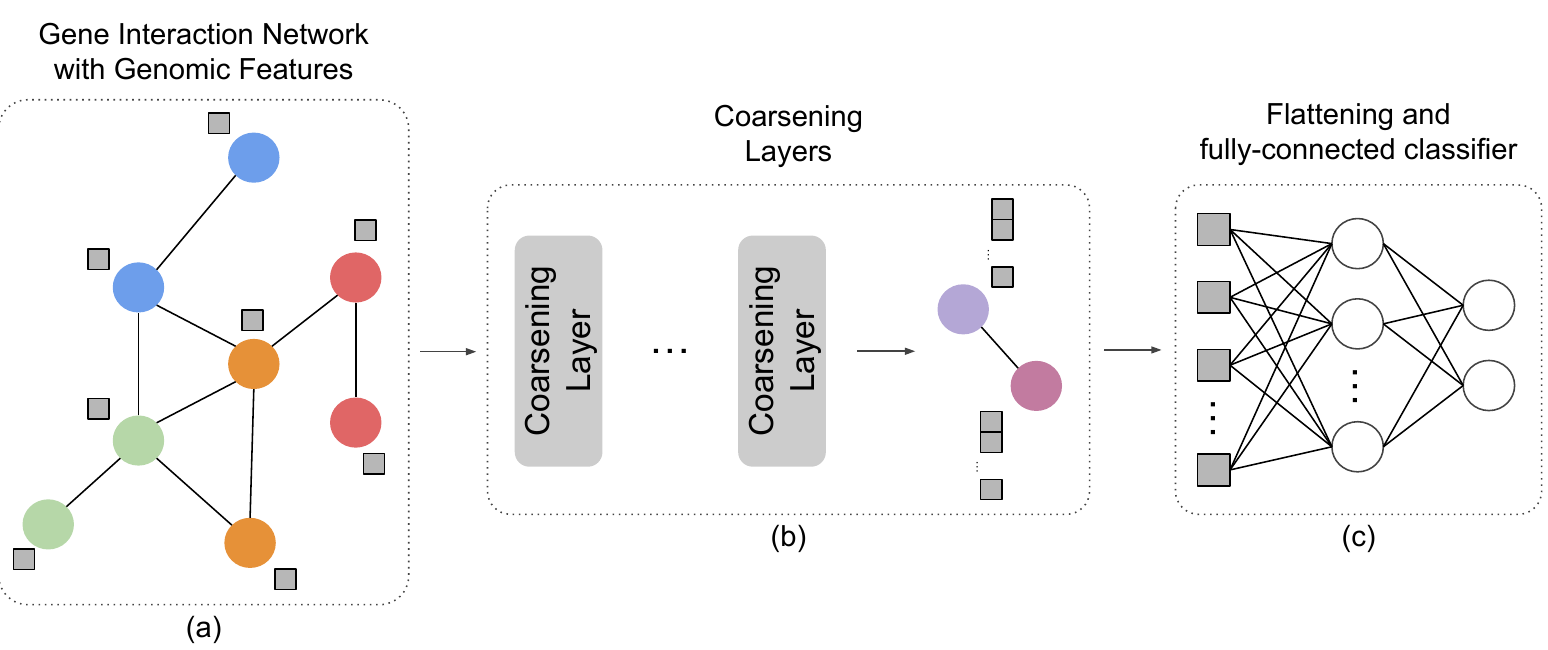}
\caption{General architecture of the GNNs used in this work. In (a), the colors represent the clusters of each protein at the input graph. These nodes undergo multiple pooling and non-linearity layers (b) and their final embeddings are concatenated and used as inputs to a FC network (c).}
\label{fig:generalModel}
\end{figure*}

\subsection{Models}\label{sec:model}
The general GNN architecture used in this work is shown in Figure \ref{fig:generalModel}. 
The model takes as inputs a gene interaction network and genomic data associated with the genes (Figure \ref{fig:generalModel}-(a)). 
We represent the genomic data associated with a node by the gray square by its side. 
In this work, we have used RNA-seq data, so that there is a single feature associated with each node -- indicated by the single squares.
However, one could also include other genomic data by concatenating their values to the previous ones.
Other works have obtained improved results by using CNV features together with RNA-seq \citep{li_cancer_2021}.

In Figure \ref{fig:generalModel}-(b), the network and the associated data goes through multiple \textit{coarsening} layers, which consist of a convolution operation, pooling and a ReLU non-linearity. 
In our experiments, we used the ChebConv  with $K=2$, that is, so that the convolutions include only the node itself and its immediate neighbors.
We found that it was easier to fit than other models and it had been previously used in related work.

The coarsening layers (Figure \ref{fig:generalModel}-(b)) aggregate the representations of nodes that belong to the same pre-computed clusters (see Section \ref{sec:clusters}), resulting in a new graph whose nodes we refer to as \textit{supernodes}.  
In this work, we used a weighted-pooling approach for aggregating the supernode representations, partly inspired by the CancelOut layer \cite{borisov2019cancelout}.

We can think of the hierarchical clusters in the network as assignment matrices $\textbf{S}^{(l)} \in \{0, 1\}^{N_l \times N_{l+1}}$, where $N_l$ and $N_{l+1}$ are the number of supernodes before and after pooling, $l = 0, ..., L$ with $L$ being the number of coarsening layers and $N_0$ the number of nodes in the original network.
The assignment matrix is given by $S^{(l)}_{ij} = 1$ if node $i$ is assigned to cluster $j$ and $0$ otherwise.
If we let $\mathbf{w}^{(l)} \in \mathbb{R}^{N_{l}}$ be a learnable weight vector and $\odot$ represent a point-wise multiplication, then the result of the weighted-pooling operation can be written as
\begin{equation}
    \mathbf{H}_{P_w}^{(l)} = (\mathbf{S}^{(l)})^\mathsf{T} ( \mathbf{w}^{(l)} \odot \mathbf{H}_{c}^{(l)}),
\end{equation}
where $\mathbf{H}_{c}^{(l)} \in \mathbb{R}^{N_l \times F_{l}}$ is the feature matrix of the nodes produced after the graph convolution, $\mathbf{H}_{P_w}^{(l)} \in \mathbb{R}^{N_{l+1} \times F_{l}}$ is the pooled value which will be used to assign the supernode value at the coarsened graph, and $\mathbf{w}^{(l)}$ is broadcast to reach the number of columns of $\mathbf{H}_{c}^{(l)}$.
Note that the weight vector can take negative values and don't necessarily need to sum to 1.

We chose a weighted pooling approach because convolutions are invariant to translations of the graph signal and apply the same filters on all the nodes. 
In images, this is generally desirable. 
However, in our context, we do not want to completely ignore the identity of each gene, since they are important by themselves. 
Therefore, we added weights to the pooling operation so that the model can learn parameters that are dependent on the genes themselves.

Finally, we used a NN containing a single hidden layer with 256 neurons as the final classifier (Figure \ref{fig:generalModel}-(c)).
We used shallow networks because they have been shown to perform well for omics data \citet{yu_architectures_2019} and in our tests we also found no gain in using deeper models.
The number of input neurons was equal to the number of features in the flattened embeddings after the coarsening layers, and when only the NN was used, without any coarsening, (e.g. as the baseline in the experiments) the number of input neurons was the same as the number of genes in the network.
After each layer besides the last one, we have used dropout, batch normalization and the ReLU as the non-linear activation function.
For the last layer of the model, however, we have either used a sigmoid or the softmax function, depending on whether it was a binary or multi-category classification task.


\subsection{Graph Clusters}\label{sec:clusters}
\begin{figure}[h]
\centering
\includegraphics[width=0.85\linewidth]{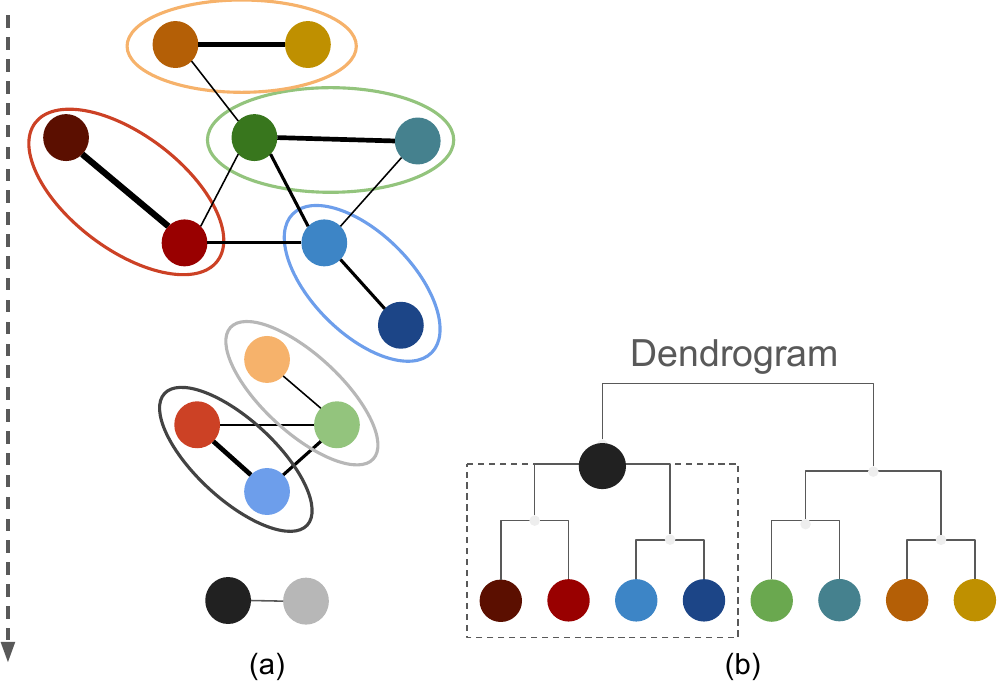}
\caption{(a) Two steps of the heavy-edge matching algorithm. Starker edges represent stronger connections and the colored ellipses indicate which nodes are aggregated together. (b) A dendrogram representing the clusters of input nodes associated with the supernodes. The supernode in black corresponds to the aggregation of the reddish and blueish inputs.}
\label{fig:hemCluster}
\end{figure}

The model in Section \ref{sec:model} requires a pre-computed hierarchical clustering of the input network.
Following similar works \citep{li_cancer_2021,wang_single-cell_2021}, we used the heavy-edge matching scheme, which consists simply of visiting the unmatched nodes in random order and matching each with the (unmatched) neighbor possessing the maximum edge weighted (Figure \ref{fig:hemCluster}-(a)).
At each match, the graph-cut of the resulting coarser graph is reduced \citep{karypis1998fast}.
More specifically, we applied the heavy edge matching algorithm to the STRING network with 14133 nodes generated as described in Section \ref{sec:dataProcessing} and obtained seven coarsened versions of the graph, each with approximately half the number of the nodes of the other, with a final graph containing 111 supernodes.
Each supernode in a coarsened graph is associated with a cluster of genes sharing a neighborhood in the STRING network.
This is illustrated in Figure \ref{fig:hemCluster}-(b), where the supernode in black and its associated input nodes are highlighted as an example.

In order to analyze the cluster of genes associated with a supernode, we performed an over-representation analysis (ORA) using the WebGestalt tool\footnote{https://www.webgestalt.org/} on each cluster.
The basic idea of an ORA is to compute how likely it was that the given set of genes would overlap with another known gene set (for example, a pathway or a  biological processes) if the genes were randomly selected from a reference set.
In our experiments, we have generally used clusters of 128 genes and a reference set containing all protein-coding genes (since we have restricted ourselves to the STRING network).
If, for instance, our set has 90 genes overlapping with a biological processes that encompasses 100 genes, then it is very unlikely that this was only due to random chance, and we can say that this biological processes is \textit{enriched}.
On the other hand, if we find that only 1 gene overlapped, then our set of genes is no more related to that biological process than a random set of genes is.

\subsection{Model Interpretation using Saliency Maps}\label{sec:methodInterpretation}
A classification model produces scores for each of the classes being classified which indicate how likely it is that the input belongs to that class.
In general, we can write $\textbf{S} = f_\theta(\textbf{x})$, where $\theta$ indicates the set of parameters, $\textbf{x}$ is the input vector and $\textbf{S} = (S_1, S_2, ..., S_C)$ is the output vector containing scores for each of the $C$ classes.
The fundamental idea of the saliency maps proposed by \citet{simonyan_deep_2013} is based on the first order approximation of the score $S_c$ produced by the model given an input $\textbf{x}_0$ for a class $c$,
\begin{equation}\label{eq:saliencyScore}
    S_c(x_0) \approx \textbf{w}^T \textbf{x}_0 + b
\end{equation}
The value of $\textbf{w}$ is then given by the gradient 
\begin{equation}
    \textbf{w} = \frac{\partial S_c}{\partial \textbf{x}}\bigg|_{x_0}
\end{equation}
One can therefore interpret the absolute values of the gradient as an indication of which input features need to change the least to produce the greatest variation in this class score.
It is reasonable then to consider those as the main features in a model's decision.
The absolute value of the gradient of a class score with respect to an input sample is referred to as the class saliencies.
\citet{mostavi2020convolutional} used saliency maps to analyze a CNN proposed for classifying the cohort-of-origin of tumor samples.
In their work, they summarized the main genes for each cohort by taking the mean value of each gene saliency over all the samples that belong to the cohort.
This implicitly assumes that the models are basing themselves on similar sets of genes for making decisions related to samples of the same cohort.
Although this is reasonable in the context of gene expression classification where one expects to find a consistent set of genes, here we have generally analyzed the distribution of saliencies across samples.

Furthermore, we have also considered the saliencies of the embeddings at the coarsest levels of the graph (for example, at the black node in Figure \ref{fig:hemCluster}).
This was partially inspired by the work in \citet{hayakawa_pathway_2022}, where they propose a model that allows interpretation at the level of the KEGG pathways.
In it, the authors use one graph neural network per pathway in the KEGG database, where each node in the KEGG networks corresponds to a gene. 
The GNNs produce embeddings that are then aggregated together in a fully connected layer and used for classification of diffuse large B-cell lymphoma subtypes.
They then show that SHAP values can be used for computing the importance of the embeddings generated by each GNN, which in turn imply in an importance of the KEGG pathway represented by the corresponding GNN.

The GNN module of our models can be seen as a learned dimensionality reduction method that transforms the dataset into a lower dimensional space composed of the embeddings of each supernode of the most coarsened level of the graph.
When we analyze these embeddings using the saliency methods as described above, we are therefore analyzing the inputs of this transformed dataset. 
Furthermore, since the graph convolutions and pooling are local operations, each supernode has a correspondence with the initial nodes (genes) that were clustered together during the pooling operations.
Hence, by analyzing the influence of the supernode on the outputs (that is, by analyzing its saliencies) we are equivalently studying the effect that the cluster of genes has, as a group, in the outputs of the model.
For instance, by analyzing the saliency of the embeddings of the black supernode in Figure \ref{fig:hemCluster}, we argue that we are implicitly analyzing the non-linearly combined effect of the colored input nodes associated with the black supernode.

Finally, since the clusters of genes associated with a supernode are generally related with biological processes (as explained in Section \ref{sec:clusters}), analyzing their effects on the output gives us an idea of how that biological process relates to the model's classifications.

\section{Experiments and Results}
Our experiments aim at (1) evaluating the effect that graph convolutions and coarsening have on the performance of the models in RNAsseq tasks (Section \ref{sec:exp1}), and (2) understanding whether these models can be interpreted to provide interesting biological insights (Section \ref{sec:exp2}).

Here, we present only the results for the tumor prediction task, but data for the cohort of origin classification can be found in the Supplementary Material.
We also focus on showing the F1-Macro average score, because it gives equal weights to both classes F1 scores, making it more robust against class imbalance.

\subsection{Pooling and Convolutions Effects on Performance}\label{sec:exp1}
\begin{figure}[h]
\centering
\includegraphics[width=0.85\linewidth]{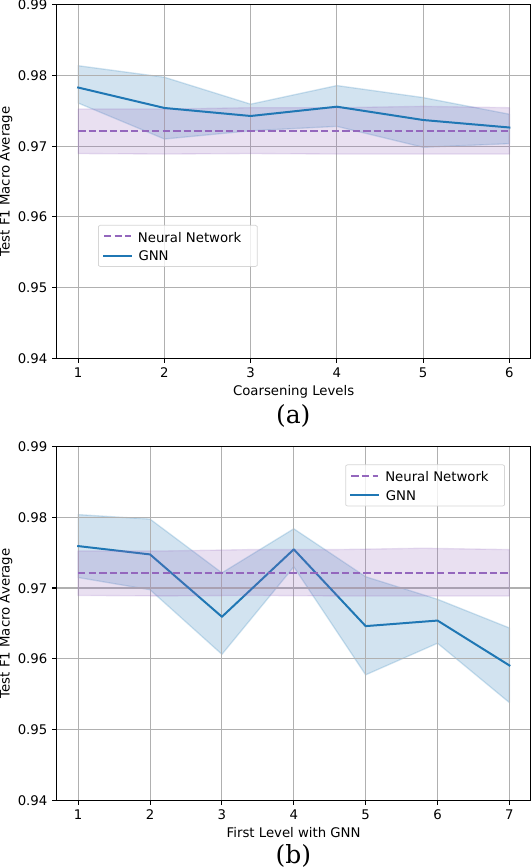}
\caption{Performance on tumor prediction task drops slightly as the number of pooling + graph convolution layers increase, with the highest score obtained with a single layer (a). Having no graph convolutions at the initial layers leads to performance drop (b).}
\label{fig:f1score}
\end{figure}

In our first experiment, we varied the number of coarsening layers in the model, going from 1 to 6.
The results are shown in Figure \ref{fig:f1score}-(a). 
The mean F1-score for a single coarsening layer was $0.978$, slightly above that of the the NN ($0.972$) but not significantly.
We also see downward trend in performance as more coarsening layers are included, a result that was also observed in the cohort classification task (Supplementary Material).

Nevertheless, the convolution operations are important to retain performance as the graph is coarsened.
Figure \ref{fig:f1score}-(b) shows the score when only the weighted pooling and non-linearities are used at the first $n$, with no convolutions. 
A similar trend is also observed in the cohort-classification task in the Supplementary Material.
This observation implies in a trade-off between performance and computational cost, since performing the graph convolutions over large networks such as ours (with over 17 thousand nodes) is quite costly both in terms of memory and computing.

The results presented here indicate that there is no performance gain obtained by increasing the number of coarsening and convolution layers.
Although different from our initial hypothesis, we observed that in most works that reported improvements due to the introduction of graph convolutions just a single convolution was used (see, for example, \citet{yin_scgraph_2022, ramirez_classification_2020, ramirez_prediction_2021}).
A possible explanation for the performance drop is the \textit{over-smoothing} \citep{rusch_survey_2023} effect, which consists of a tendency of node features to become more similar as the depth of the GNN increases, but we leave this exploration for future work.

\subsection{Model Interpretation}\label{sec:exp2}
In this section, we trained and explained the decisions of one of the models using the architecture with seven coarsening levels, weighted pooling and convolutions starting at the fifth level.

This architecture was chosen because the supernodes at the final coarsening levels were associated with 128 genes, which is a good number of genes to perform over representation analysis (ORA), while retaining sufficient performance.
Furthermore, it represents a significant reduction in the dimensionality of the original network without the costly graph convolutions at the more detailed levels of the graphs.
Specifically, the nodes at the fifth coarsening level (the first where a convolution is performed in this model) correspond to clusters of 16 genes, aggregated using only weighted pooling and non-linearities.
Since there are only three convolutional layers, the model is more easily interpretable as well, since it induces 8-dimensional embeddings at each node of the coarsest graph (at each convolutional layer, we double the number of nodes in the embeddings).
Using the model possessing seven coarsening levels, all with convolutions, would result in 32 dimensions per node in our case, which could complicate the analysis.
Finally, it is worth noting that in comparison with the baseline NN, this architecture has less than 10\% of the number of learnable parameters.

\begin{figure*}[!h]
    \begin{center}
        \includegraphics[width=\linewidth]{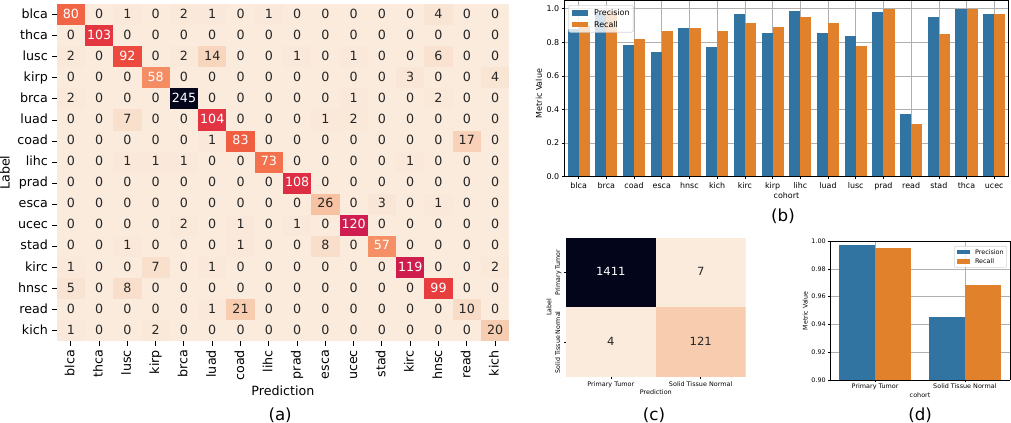}
    \end{center}
    \caption{Test results on cohort classification and the tumor prediction tasks using a multi-task model. The confusion matrix for the Pan-cancer Cohort classification is shown in (a) together with the precision and recall obtained for each category in (b), whereas the confusion matrices and precision-recall plots are shown in (c) and (d) for the Pan-cancer Tumor Prediction task.}
    \label{fig:cohortAndTypePerformance}
\end{figure*}

We show in Figure \ref{fig:cohortAndTypePerformance} confusion matrices for cohort classification and tumor prediction along with relevant metrics.
One can see that the model is able to achieve almost perfect separation among tumorous and non-tumorous samples, with an accuracy over all samples, including all cohorts, of $99.3\%$, which is on the level of other state-of-the-art works.
The model is also able to distinguish appropriately between most of the cohorts, but performs poorly for some of them (Figures \ref{fig:cohortAndTypePerformance}-(a) and \ref{fig:cohortAndTypePerformance}-(b)).
In particular, most of the error is in distinguishing samples for the READ and COAD cohorts.
This is also the case in other works, such as \citet{mostavi2020convolutional}.
One could argue that this is due to the lack of READ samples, however, KICH and ESCA cohorts also possess a similar amount of examples and show better recall and precision.
A more likely explanation is that these cancers share similar processes due to their anatomical proximity.

\subsubsection{Gene Biomarkers}
Next, we consider here the \lq importance\rq\ assigned by the model to each of the input genes in the Tumor Prediction task.
To do that, we computed the saliencies of the input genes, as explained in Section \ref{sec:methodInterpretation}.
Analyzing the top ranked genes, we find various known cancer-related genes.
The top 10 genes are summarized in Table \ref{table:genesTypes}. 
\begin{table}[h]
    \caption{Main genes obtained through the saliency analysis of the multi-label model for Primary Tumor vs Normal Tissue classification.}
    \begin{tabular}{@{}lll@{}}
        \toprule
        \multicolumn{1}{c}{Rank} & \multicolumn{1}{c}{\begin{tabular}[c]{@{}c@{}}STRING \\ Protein Id\end{tabular}} & \multicolumn{1}{c}{Gene Symbol} \\ \midrule
        1                        & ENSP00000436785                                                                  & \textit{SLC35F2}                         \\
        2                        & ENSP00000325808                                                                  & \textit{LRRN4CL}                         \\
        3                        & ENSP00000416508                                                                  & \textit{ATP13A3}                         \\
        4                        & ENSP00000309432                                                                  & \textit{RETREG3}                         \\
        5                        & ENSP00000360540                                                                  & \textit{CEP55}                           \\
        6                        & ENSP00000428263                                                                  & \textit{DGLUCY}                          \\
        7                        & ENSP00000437550                                                                  & \textit{LATS1}                           \\
        8                        & ENSP00000272348                                                                  & \textit{SNRPG}                           \\
        9                        & ENSP00000344456                                                                  & \textit{CTNNB1}                          \\
        10                       & ENSP00000478783                                                                  & \textit{LTN1}                          \\ \bottomrule
    \end{tabular}
    \label{table:genesTypes}
\end{table}

By searching in the literature, we found all of these genes to have been previously studied in the context of cancer in various forms.
More interestingly, five of them were associated with more than one type of cancer, indicating that the pan-cancer tumor prediction model is at some level aiming at general cancer genes, instead of building a list of cohort-specific biomarkers.
One of these five genes is \textit{SLC35F2}, which was found to be highly expressed in various human cancers \citep{winter2014solute}.
\textit{LATS1}, which encodes the Large Tumor Suppressor Kinase 1, was also associated with multiple tumors and is related with cancer cell growth \citep{pan2019cell}.
\textit{SNRPG} also plays a role in tumor development and has been associated with various human cancers.
Furthermore, it has been suggested to have potential for oncogenic drug discovery \citep{mabonga2019oncogenic}.
Lastly, mutations in the \textit{CTNNB1} genes are related to multiple cancer types \citep{gao_exon_2018} as well and \textit{CEP55} over-expression was previously correlated with poor prognosis of various tumor types \citep{jeffery2016beyond}.

We found the other genes to be associated with cohort-specific cancers.
One of these, \textit{LRRN4CL}, was recently associated with pulmonary metastasis in mice and correlates with decreased survival of melanoma patients \citep{van2021crispr}.
\textit{ATP13A3} has been considered as a biomarker for pancreatic cancer therapies \citep{madan2016atp13a3} and is associated with a decreased overall survival in pancreatic cancer \citep{sekhar_atp13a3_2022}.
\textit{RETREG3} (\textit{FAM134C}) is a member of the \textit{FAM134} family \citep{reggio_role_2021}. 
One of the members of this family, \textit{FAM134B}, has been studied in the context of colorectal cancer \citep{kasem2014roles}.
Finally, \textit{DGLUCY} is related with progression of gastric cancer \citep{zhu_c14orf159_2019} and \textit{LTN1} was linked with ovarian cancer prognosis.

\subsubsection{Supernode Interpretation}
\begin{figure}[h]
    \centering
    \includegraphics[width=\linewidth]{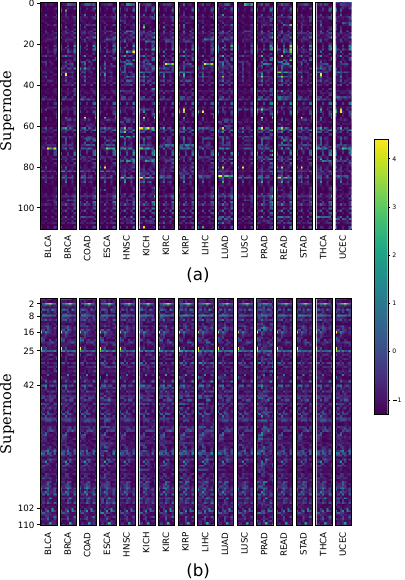}
    \caption{Example embeddings (a) and their saliencies (b) produced when distinguishing normal from tumorous samples. Each row corresponds to a supernode in a coarsened version of the STRING.}
    \label{fig:typeSalienciesMap}
\end{figure}

We consider now the interpretation of the model supernodes in the Tumor Prediction task.
Recall that the output of the coarsening layers of the model considered here consists in 8-dimensional embeddings, one for each of the 111 supernodes in the coarsest graph.
Furthermore, each supernode is primarily associated with the cluster of protein-encoding genes in the original STRING network that were aggregated by the coarsening layers, until arriving at the particular supernode.
We computed the saliencies for each of the embeddings in the same way as done with the input genes, which resulted in a $111 \times 8$ matrix of supernode saliencies.
Figures \ref{fig:typeSalienciesMap}-(a) and \ref{fig:typeSalienciesMap}-(b) show a set of examples of the resulting supernode embeddings and their corresponding saliencies for the Tumor Prediction task.
We randomly selected one example from each cohort.

\begin{figure}[h]
    \centering
    \includegraphics[width=\linewidth]{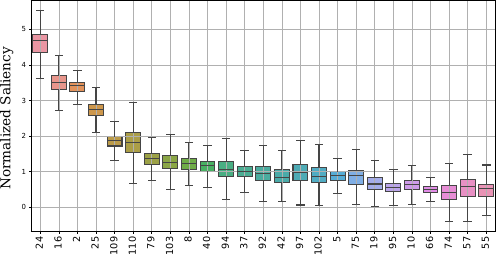}
    \caption{The distribution of each supernode saliency over the dataset examples.}
    \label{fig:typeSalienciesBox}
\end{figure}

It is interesting to note that, independently of the cohort, the distinction between tumorous and normal samples are always related to a small set of supernodes, even though the embeddings differ depending on the cohort.
Figure \ref{fig:typeSalienciesBox} further confirms this, where we show the distribution of the normalized values of the saliencies.
Specifically, for each example individually, we normalized the values of the 128 saliencies (one for each supernode) to zero mean and unit standard deviation, so that their values are comparable across the examples.
If the model were using different processes to predict tumor samples coming from different cohorts, then we would expect that the saliencies would show a multimodal distribution or at least be more spread over the importance range.
However, we see that the top-ranked supernodes are consistently in the top.

The question remains to whether these supernodes are related to meaningful sets of genes.
To evaluate this, we performed an over-representation analysis using WebGestalt\footnote{https://www.webgestalt.org/} on the clusters of input genes associated with the top-4 supernodes.
The results, presented in Table \ref{table:oraTypes}, show that the supernodes are enriched with respect to a few biological processes, some of which have been studied with relation to various cancer types.
The mitochondrial respiratory chain complex assembly gene set, associated with cluster 25, for example, has been studied in the context of cancer aggressiveness \citep{simonnet2002low} and tumorigenesis \citep{lemarie2011mitochondrial}.
\begin{table*}[h]
    \caption{Gene sets obtained from the over-representation analysis performed on the gene clusters associated with the supernodes selected through their saliencies on the test data.}
    \begin{center}
        \begin{tabular}{@{}cllll@{}}
            \toprule
            \multicolumn{1}{l}{\textbf{Supernode}} & \textbf{Gene Set} & \textbf{Description}                                  & \textbf{\begin{tabular}[c]{@{}l@{}}Enrichment \\ Ratio\end{tabular}} & \textbf{FDR}       \\ \midrule
            \multirow{4}{*}{24}                  & GO:0008202        & steroid metabolic process                             & 27.130                                                               & \textless{}2.2e-16 \\
                                                   & GO:0006720        & isoprenoid metabolic process                          & 20.738                                                               & \textless{}2.2e-16 \\
                                                   & GO:0062012        & \makecell[l]{regulation of small molecule \\ metabolic process}        & 8.9674                                                               & 8.9462e-12         \\
                                                   & GO:0042737        & drug catabolic process                                & 11.341                                                               & 2.1444e-6          \\ \midrule
            \multirow{3}{*}{16}                  & GO:0006022        & aminoglycan metabolic process                         & 34.386                                                               & \textless{}2.2e-16 \\
                                                   & GO:0009100        & glycoprotein metabolic process                        & 20.225                                                               & \textless{}2.2e-16 \\
                                                   & GO:1903509        & liposaccharide metabolic process                      & 14.361                                                               & 2.7721e-8          \\ \midrule
            \multirow{2}{*}{2}                   & GO:0034453        & microtubule anchoring                                 & 62.127                                                               & 1.2218e-7          \\
                                                   & GO:0044839        & cell cycle G2/M phase transition                      & 7.5836                                                               & 0.018845           \\ \midrule

            \multirow{5}{*}{25}                  & GO:0010257        & \makecell[l]{NADH dehydrogenase complex \\ assembly}                   & 73.730                                                               & \textless{}2.2e-16 \\
                                                   & GO:0033108        & \makecell[l]{mitochondrial respiratory chain \\ complex assembly}      & 49.153                                                               & \textless{}2.2e-16 \\
                                                   & GO:0009141        & \makecell[l]{nucleoside triphosphate metabolic \\ process}             & 19.269                                                               & \textless{}2.2e-16 \\
                                                   & GO:1902600        & proton transmembrane transport                        & 18.420                                                               & \textless{}2.2e-16 \\
                                                   & GO:0099132        & \makecell[l]{ATP hydrolysis coupled \\cation transmembrane transport} & 24.821                                                               & 8.6641e-9          \\ \bottomrule
            \end{tabular}
    \end{center}
    \label{table:oraTypes}
\end{table*}
With respect to supernode 16, the cluster associated with it was significantly over-represented in the glycoprotein metabolic process, which is defined in GeneOntology (GO) as all the pathways and chemical reactions involving glycoproteins\footnote{https://www.ebi.ac.uk/QuickGO/term/GO:0009100}.
Searching the literature, we found the work by \citet{kailemia_glycans_2017}, which reviewed various glycoproteins biomarkers that are used for monitoring and screening patients with a wide sort of cancer types.
Supernode 2 was also associated with interesting processes, such as microtubule anchoring.
The GO entry for microtubule anchoring states that it encompasses any process where a microtubule is maintained in a specific cell locations \footnote{https://www.ebi.ac.uk/QuickGO/term/GO:0034453}.
We found that there is indeed evidence suggesting the existence of dysfunctions in microtubule-related processes in cancer cells \citep{draber_dysregulation_2021}.
Interestingly, microtubules have also been targeted for anticancer therapies for decades \citep{dumontet_microtubule-binding_2010}.
Finally, the gene set relating to the isoprenoid metabolic process was over-represented in the genes associated with supernode 24.
According to the GO entry for the process, it comprises the chemical processes and pathways involving isoprenoid components \footnote{https://www.ebi.ac.uk/QuickGO/term/GO:0006720}.
Following \citet{wiemer_intermediate_2009}, isoprenoid biosynthesis is related to cancer cell growth and metastasis, making it a target for anticancer therapies.

\section{Discussion and Limitations}
The performance results in Section \ref{sec:exp1} showed a drop in performance with an increase in the number of layers.
We experimented our methodology with other networks, such as using similarities graphs, but not systematically.
It possible therefore that other biological networks would show improved results.
\citet{yin_scgraph_2022}, for example, when working with single-cell RNA-seq data, found that the best results were actually obtained when only the top-1\% strongest edges were kept in the STRING network.
When we considered that, we observed that such reduction in the network created a significant number of singleton nodes.
In the context of a single pooling layer, this results in just a few actual convolutions being applied, since the singleton nodes do not have neighbors.
Furthermore, applying hierarchical clustering algorithms over graphs with singleton nodes will in general lead to the existence of graph clusters containing only the singleton node, which would make it impossible to perform an over-representation analysis.

The pooling approach used here is somewhat restrictive, as it employs a fixed hierarchical cluster structure computed prior to model training. 
The benefit of this approach is that it is computationally tractable, which is important in our context, as we deal with networks that have tens of thousands of nodes.

However, it is possible that trainable graph pooling approaches that allowed the pre-computed hierarchical structure to be adapted during training could lead to different and perhaps improved results.
Finally, this work has focused on the ChebConv, as have some of the other works in the area.
We chose this network because we had difficult in fitting other architectures such as the GCN \citep{welling_semi-supervised_2016} and the GraphSAGE \citep{hamilton2017inductive}, and the available Graph Attention Network (GAT) \citep{velivckovic2017graph} implementations could not be easily integrated in our pipeline.
Recently, also, transformer architectures have achieved achieved prominence and also explored in the context of graph ML, as in the Graph Transformer Network \citep{yun_graph_nodate}, but we did not consider them here.
It is not unreasonable, therefore, to expect that different choices of architectures would influence the results.

The model interpretation resulted in various widely known cancer-related genes, and we were also able to find interesting biological processes that have been studied in the context of cancer and even considered for the development of treatments.
Although this is encouraging, we need to consider some limitations that we hope to tackle in future works.

First of all, when we interpret the supernodes, we generally find more than one significantly over-represented gene sets.
However, it is not clear which of them, or if all, is the most important, and it is also not clear whether it is just a smaller set of genes in the cluster that is relevant, instead of the biological process itself.
One possible solution to build models that could overcome this is to introduce learnable clustering functions instead of the fixed, pre-computed, hierarchical clusters as we have done.
In doing that, we would hope that the learned graph clusterings would include reduce to a single biological meaningful pathway for the task.

Also related to the interpretation of the supernodes when convolutions are present, we note that, even though the graph convolutions are designed as local operations, modeling only interactions between neighboring nodes, they lead to interactions between nodes that will not necessarily be pooled together in the same final supernode.
Therefore, another interesting question that emerges is whether the attributed importance of a supernode is due to the genes belonging to the associated cluster itself, or if it is more related to how this cluster of genes is interacting with its neighbors.

One way in we which we consider deepening our analysis in this sense is in introducing learnable edge weights, as was already done by \citet{yin_scgraph_2022}.
In doing that, we hope that the edge weights associated with a supernode will diminish towards 0, if the supernode is relevant only through the genes it encompasses and the biological process it represents.
Similarly, if it is the interactions with particular neighbors that are important, then the edge weights would be non-zero.
This behaviour of the weights could be encouraged through regularization.

It could also be interesting to experiment with other techniques for model explanation such as Grad-cam \citep{selvaraju2017grad}, which uses the final convolution layer to produce a coarse map over the original input -- although they would likely require adaptations for graph data.

\section{Conclusion}
In this work, we have considered the effects that multiple coarsening levels had on performance and whether different forms of applying pooling and the use of convolutions would alter the results.
In general, we have observed that coarsening the input graph for more than one layer either reduced or just maintained the performance, in comparison to a NN.

We have also explored the use of saliency analysis to interpret our studied GNNs.
Besides using these methods to extract biomarkers at the gene level, as was already done in previous works using NNs \citep{mostavi2020convolutional}, we have also considered the saliencies of the supernodes, generated through the pooling steps performed over a pre-computed hierarchical cluster structure.
In both cases, we obtained interesting biomarkers and biological processes, many of which were previously studied in relation with cancer.

\backmatter

\bmhead{Supplementary information}



\bmhead{Acknowledgements}
Some experiments in this work used the PCAD infrastructure, http://gppd-hpc.inf.ufrgs.br, at INF/UFRGS.


\section*{Declarations}




\bmhead{Funding}
This study was financed in part by the Coordenação de Aperfeiçoamento de Pessoal de Nível Superior - Brasil (CAPES) - Finance Code 001, and by grants from the Fundação de Amparo à Pesquisa do Estado do Rio Grande do Sul - FAPERGS [21/2551-0002052-0 and 22/2551-0000390-7 (Project CIARS)] and Conselho Nacional de Desenvolvimento Científico e Tecnológico (CNPq) [308075/2021-8].

\bmhead{Conflict of Interest} The authors have no competing interests to declare that are relevant to the content of this article.

\bmhead{Ethics approval and consent to participate} Not applicable.

\bmhead{Consent for publication} Not applicable.


\bmhead{Author contribution} Conceptualization: Thomas V. Fontanari, Mariana Recamonde-Mendoza; Methodology: Thomas V. Fontanari, Mariana Recamonde-Mendoza; Formal analysis and investigation: Thomas V. Fontanari, Mariana Recamonde-Mendoza; Writing - original draft preparation: Thomas V. Fontanari; Writing - review and editing: Mariana Recamonde-Mendoza; Funding acquisition: Mariana Recamonde-Mendoza; Resources: Thomas V. Fontanari, Mariana Recamonde-Mendoza; Supervision: Mariana Recamonde-Mendoza











\bibliography{sn-bibliography}

\end{document}